\documentclass{article} 
\usepackage{nips15submit_e,times}
\usepackage{hyperref}
\usepackage{url}
\usepackage{amsmath}
\usepackage{algorithm}
\usepackage{algpseudocode}
\usepackage[psamsfonts]{amssymb}
\usepackage[pdftex]{graphicx}

\newcommand{\av}{\mathbf{a}}

\newcommand{\kv}{\mathbf{k}}

\newcommand{\x}{\mathbf{x}}
\newcommand{\y}{\mathbf{y}}

\newcommand{\I}{\mathbf{I}}
\newcommand{\K}{\mathbf{K}}

\newcommand{\X}{\mathbf{X}}

\newcommand{\N}{\mathcal{N}}

\title{Local Nonstationarity for Efficient Bayesian Optimization}

\author{
Ruben Martinez-Cantin
\\
Centro Universitario de la Defensa\\
Zaragoza, 50090, Spain\\
\texttt{rmcantin@unizar.es} \\
}

\nipsfinalcopy 

\begin{document}

\maketitle

\begin{abstract}
Bayesian optimization has shown to be a fundamental global optimization algorithm in many applications: ranging from automatic machine learning, robotics, reinforcement learning, experimental design, simulations, etc. The most popular and effective Bayesian optimization relies on a surrogate model in the form of a Gaussian process due to its flexibility to represent a prior over function. However, many algorithms and setups relies on the stationarity assumption of the Gaussian process. In this paper, we present a novel nonstationary strategy for Bayesian optimization that is able to outperform the state of the art in Bayesian optimization both in stationary and nonstationary problems.
\end{abstract}

\section{Introduction}

Many problems in engineering, computer science, economics, etc., require to find the extremum of a real valued function. These functions have typically nice properties from a numerical point of view, like being continuous and sometimes smooth (e.g.: Lipschitz continuous). However, many of those functions represent costly processes, expensive trials or several time consuming computations. Furthermore, those functions might not have a closed-form expression or might be highly multimodal.

Bayesian optimization, although being a classic method \cite{Mockus1989}, has become quite popular recently for being a sample efficient method of global optimization \cite{Jones:1998}. Recent works have found connections with Bayesian optimization and the way biological systems adapt and search, such as human active search \cite{Borji2013} or animal adaptation to injuries \cite{Cully2015}. In machine learning, it has been applied for automatic algorithm tuning \cite{Snoek2012} and reinforcement learning \cite{MartinezCantin09AR}. It is specially suitable for these kind of expensive black-box functions and trial-and-error methodologies for using a Bayesian surrogate model, that is, a distribution over target functions $P(f)$. This surrogate model has a twofold purpose:
\begin{itemize}
\item It can be updated recursively as outcomes are available from the evaluated trials $\y_i = f(\x_i)$ 
\begin{equation}
  \label{eq:bayes-sec}
  P(f|\x_{1:i},\y_{1:i}) = \frac{P(\x_{i},\y_{i}|f) P(f|\x_{1:i-1},\y_{1:i-1})}{P(\x_{i},\y_{i})}, \qquad \forall \; i=2 \ldots n
\end{equation}
\item It allows a best response analysis for the decision/action $\av$ of selecting the next trial $\x_{n+1}$:
\begin{equation}
\av^{BO} = \arg \min_{\av} \int_F \delta_n(f,\av) \; dP(f|\x_{1:n},\y_{1:n})     
\label{eq:bayes-average}
\end{equation}
\end{itemize}
 where $\delta_n(f,\av)$ is the optimality criteria of regret function that drives the optimization towards the optimum $\x^*$, such as the \emph{optimality gap} $\delta_n(f,\av) = f\left(\x_n\right) - f(\x^*)$, the \emph{Euclidean distance error} $\delta_n(f,\av) = \|\x_n - \x^*\|_2$ or the \emph{relative entropy} $\delta_n(f,\av) = H(\x^*|\x_{1:n-1}) - H(\x^*|\x_{1:n})$. The first condition allows to make the decisions of the second condition with all the information available at that time, thus improving the search of the optimum.

Without loss of generality, consider the problem of finding the minimum of an unknown real valued function $f:\mathbb{X} \rightarrow \mathbb{R}$, where $\mathbb{X}$ is a compact space, $\mathbb{X} \subset \mathbb{R}^d, d \geq 1$. For the remainder of the paper, we are going to assume that the surrogate model $P(f)$ is a Gaussian process $\xi(\x)$with inputs $\x \in \mathbb{X}$ and an associate kernel or covariance function $k(\cdot,\cdot)$. One advantage of using Gaussian processes (GPs) as a prior distributions over functions is that new observations of the target function $(x_i,y_i)$ can be easily used to update the distribution over functions. Furthermore, the posterior distribution is also a GP: $\xi_i = \left[ \xi(\cdot) | x_{1:i},y_{1:i} \right]$. Therefore, the posterior can be used as an informative prior for the next iteration in a recursive algorithm.

The following algorithm summarizes the steps in Bayesian optimization. Note that, without loss of generality, for the remainder of the paper we are going to assume a Gaussian process with MCMC on the hyperparameters as surrogate model and the expected improvement as the acquisition function. However, the proposed algorithms also work with other popular models such as Student-t processes \cite{O'Hagan1992,AmarShah2014}, a discrete representation of the hyperparameters \cite{krause07nonmyopic} or different acquisition functions such as upper confidence bound \cite{Srinivas10} or relative entropy \cite{NIPS2014_5324,HennigSchuler2012}, among others.

\begin{algorithm}
\caption{Bayesian optimization (BO) with MCMC}\label{al:bo}
\begin{algorithmic}[1]
\For{$n = 1 \ldots N$}
   \State $\X \gets \x_{1:n-1} \;\;\;  \y \gets y_{1:n-1}$ 
   \For{$i = 1 \ldots m$}              \Comment{We have $m$ MCMC samples}
   \State $\mu_i \gets \kv(\x,\X| \theta_i) \K(\X,\X | \theta_i) \y$  \Comment{Predicted mean}
   \State $\sigma_i \gets k(\x,\x| \theta_i) - \kv(\x,\X| \theta_i) \K(\X,\X | \theta_i) \kv(\X,\x| \theta_i)$  \Comment{Predicted variance}
   \EndFor
   \State $\Theta \gets \{\theta_i\}_{i=1}^{m}$  \Comment{Update the hyperparameters using slice sampling}
   \State $\x_n = \arg \max_{\x_{n}} \; \sum_{i=1}^m \left[\left(\y_{best} - \mu_i\right) \Phi(z_i) + \sigma_i \phi(z_i)\right]$ \label{al:crit} \Comment{Expected improvement}
   \State $\y_n \gets f(\x_n)$
\EndFor
\end{algorithmic}
\end{algorithm}

The main contribution of the paper is an algorithm for improved Bayesian optimization using a combination of local and global kernels to achieve a nonstationary behavior, called \emph{Spartan Bayesian Optimization}. Although it reaches its best performance in problems that are intrinsically nonstationary, our evaluation shows that it can improve the results of Bayesian optimization in many setups. Additionally, we also present a method to actively learn the Gaussian process hyperparameters while conducting Bayesian optimization with two alternatives: using explicit exploration driven by information gain and using simultaneous perturbations to numerically estimate the variability of the model. Finally, we add some comments about using hierarchical Bayesian optimization to deal with complex input spaces.

\section{Nonstationarity in Gaussian processes}

Many applications of Gaussian process regression, including Bayesian optimization, are based on the assumption that the process is stationary and, in many times, isotropic. For example, the use of the isotropic squared exponential kernel in GPs is quite frequent: $k_{SE}(\x,\x') = \exp((\x-\x')^T \Lambda (\x-\x'))$, where $\Lambda = \theta_l^{-1} \I$ and $\theta_l$ represents the length-scale hyperparameter that captures the smoothness or variability of the function. That is, small values of $\theta_l$ will be more suitable to capture signals with high frequency components; while large values of $\theta_l$ result in model for low frequency signals or flat functions. 

This property results in an interesting behavior in Bayesian optimization. For the same distance between points, a kernel with smaller length-scale will result in higher variance, therefore the exploration will be more aggressive. This idea has been explored previously in \cite{ZiyuWang2013} by forcing smaller scale parameters to improve the exploration. More formally, in order to achieve no-regret convergence to the minimum, the target function must be an element of the reproducing kernel Hilbert space (RKHS) characterized by the kernel $k(\cdot,\cdot)$ \cite{Bull2011,Srinivas10}. For a set of kernels like the SE or Mat{\'e}rn, it can be shown that, given two kernels $k_l$ and $k_s$ with large and small length scale hyperparameters respectively, \emph{any function $f$ in the RKHS characterized by a kernel $k_l$ is also an element of the RKHS characterized by $k_s$} \cite{ZiyuWang2013}. Thus, using $k_s$ instead of $k_l$ is safer in terms of guaranteeing convergence. However, if the small kernel is used everywhere, it might result in unnecessary sampling of smooth areas.

\begin{figure}
  \centering
  \includegraphics[width=0.3\linewidth]{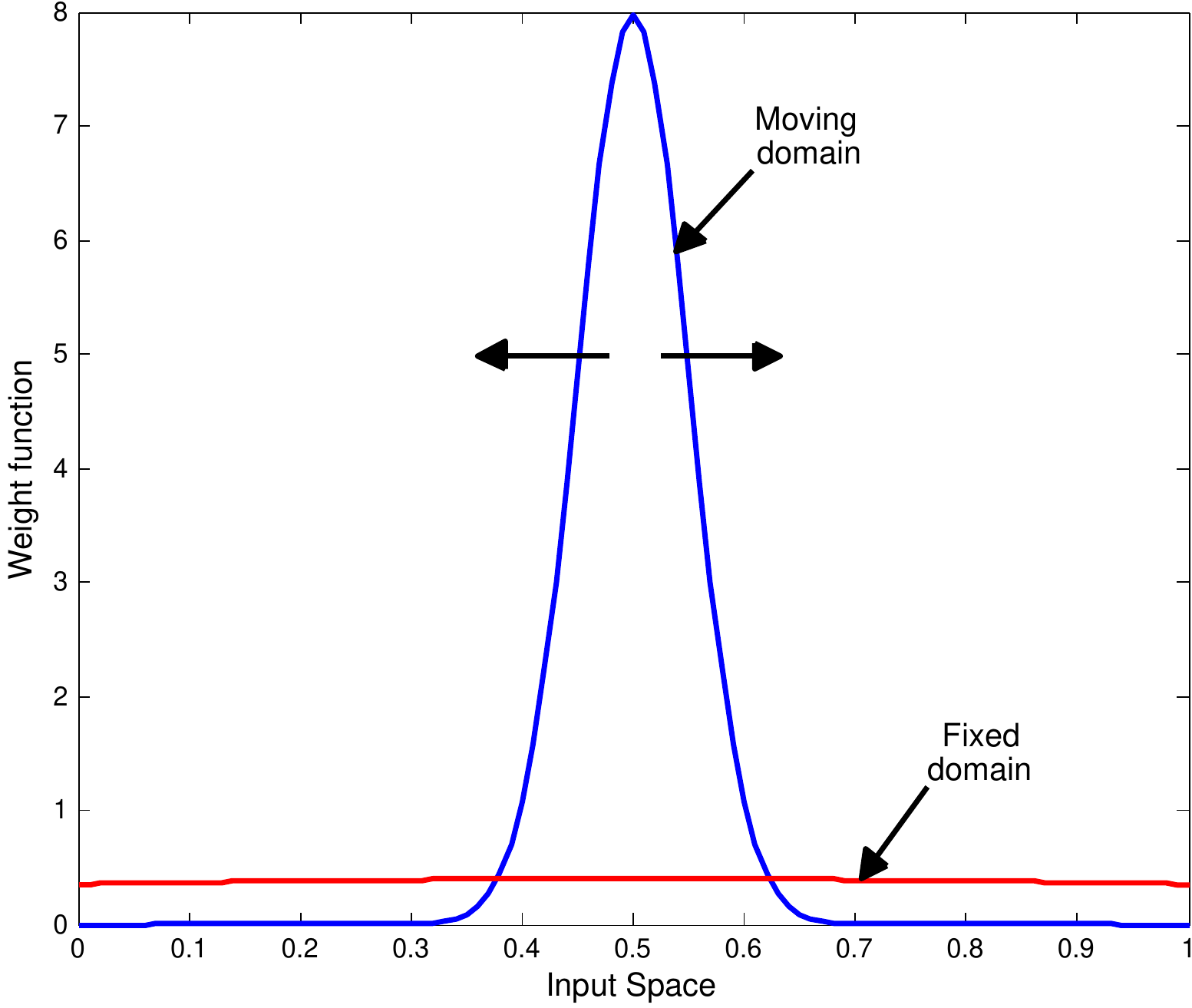}
  \caption{Weighting function of the local and global kernels for nonstationary Bayesian optimization. The global kernel (in red) guarantees full support over the whole input space. The weight of local kernel (in blue) is based on a narrow Gaussian which can be moved through the space, therefore focusing on the region near the minimum value.}
  \label{fig:domains}
\end{figure}

Nevertheless, most of the applications where Bayesian optimization is used are heterotropic and nonstationary. For example, we may have an heteroscedastic function with different variability, frequency or smoothness in different regions. Thus, these functions require kernels with different length scales for those regions. Take for example a reinforcement learning problem where many policies might result in a failure condition, returning a similar null reward. Therefore, the reward function is almost flat except for a set of parameters where it actually varies.

There has been several attempts to model nonstationary functions with Gaussian processes. The most popular is the use of specific nonstationary kernels \cite{Rasmussen:2006}, Bayesian treed GP models \cite{gramacy2005bayesian} or projecting the input space to a stationary latent space \cite{sampson1992nonparametric}. Recently, a version of the later idea has been applied to Bayesian optimization \cite{snoek-etal-2014a}.

Our approach to nonstationarity, the \emph{Spartan Bayesian Optimization} algorithm, is based on the model presented in \cite{krause07nonmyopic} where the input space is partitioned in different regions such as the resulting GP is the linear combination of local GPs: $\xi(\x) = \sum_i \lambda_i (\x) \xi_i(\x)$. Each local GP has its own specific hyperparameters, making the final GP nonstationary even when the local GPs are stationary. In order to achieve smooth interpolation between regions, Krause and Guestrin \cite{krause07nonmyopic} 
suggest the use of a weighting function $\nu_i(\x)$ for each region, having the maximum in region $i$ and decreasing its value with distance to region $i$. Then, we can set $\lambda_i(\x) = \sqrt{\frac{\nu_i(\x)}{\sum_j \nu_j(\x)}}$.

For Bayesian optimization, we suggest the combination of a local and global kernel with multivariate normal distributions as weighting functions as presented in Figure \ref{fig:domains}. Assuming that the input space is bounded to the $[0,1]^d$ hypercube, which can be achieved by rescaling the original problem, we consider that for each dimension:
\begin{equation}
  \label{eq:weights}
    \nu_{global}^{(k)} = \N(0.5, 10); \qquad   \nu_{local}^{(k)} = \N(\theta_{pos}^{(k)}, 0.05) \qquad \forall \;\; k = 1\ldots d
\end{equation}
where $\{\theta_{pos}^{(k)}\}_{1}^{d}$ is considered to part of the set of hyperparameters of the surrogate model that are learned accordingly when new data is available $\Theta = \{\theta_{pos}, \theta_{local}, \theta_{global}\}$. In that way, the position of the local kernel is adapting towards to the area near the minimum or other important area.

\begin{algorithm}
\caption{Spartan Bayesian Optimization (SBO) with MCMC}\label{al:sbo}
\begin{algorithmic}[1]
\For{$n = 1 \ldots N$}
   \State $\X \gets \x_{1:n-1} ; \;\;  \y \gets y_{1:n-1};  \;\; $
   \For{$i = 1 \ldots m$}              \Comment{We have $m$ MCMC samples}
   \State $k(\x,\x'| \theta_i) \gets \lambda_{l}(\x| \theta_i^{pos}) \lambda_{l}(\x'| \theta_i^{pos}) k_{l}(\x,\x'| \theta^l_i) + \lambda_{g}(\x) \lambda_{g}(\x') k_{g}(\x,\x'| \theta^g_i)$ 
   \State $\mu_i \gets \kv(\x,\X| \theta_i) \K(\X,\X | \theta_i) \y$  \Comment{Predicted mean}
   \State $\sigma_i \gets k(\x,\x| \theta_i) - \kv(\x,\X| \theta_i) \K(\X,\X | \theta_i) \kv(\X,\x| \theta_i)$  \Comment{Predicted variance}
   \EndFor
   \State $\Theta \gets \{\theta^{(i)}_{global}, \theta^{(i)}_{local}, \theta^{(i)}_{pos}\}_{i=1}^{m}$  \Comment{Update the hyperparameters using slice sampling}
   \State $\x_n = \arg \max_{\x_{n}} \; \sum_{i=1}^m \left[\left(\y_{best} - \mu_i\right) \Phi(z_i) + \sigma_i \phi(z_i)\right]$ \Comment{Expected improvement}
   \State $\y_n \gets f(\x_n)$
\EndFor
\end{algorithmic}
\end{algorithm}

\begin{figure}
  \centering
  \includegraphics[width=0.28\linewidth]{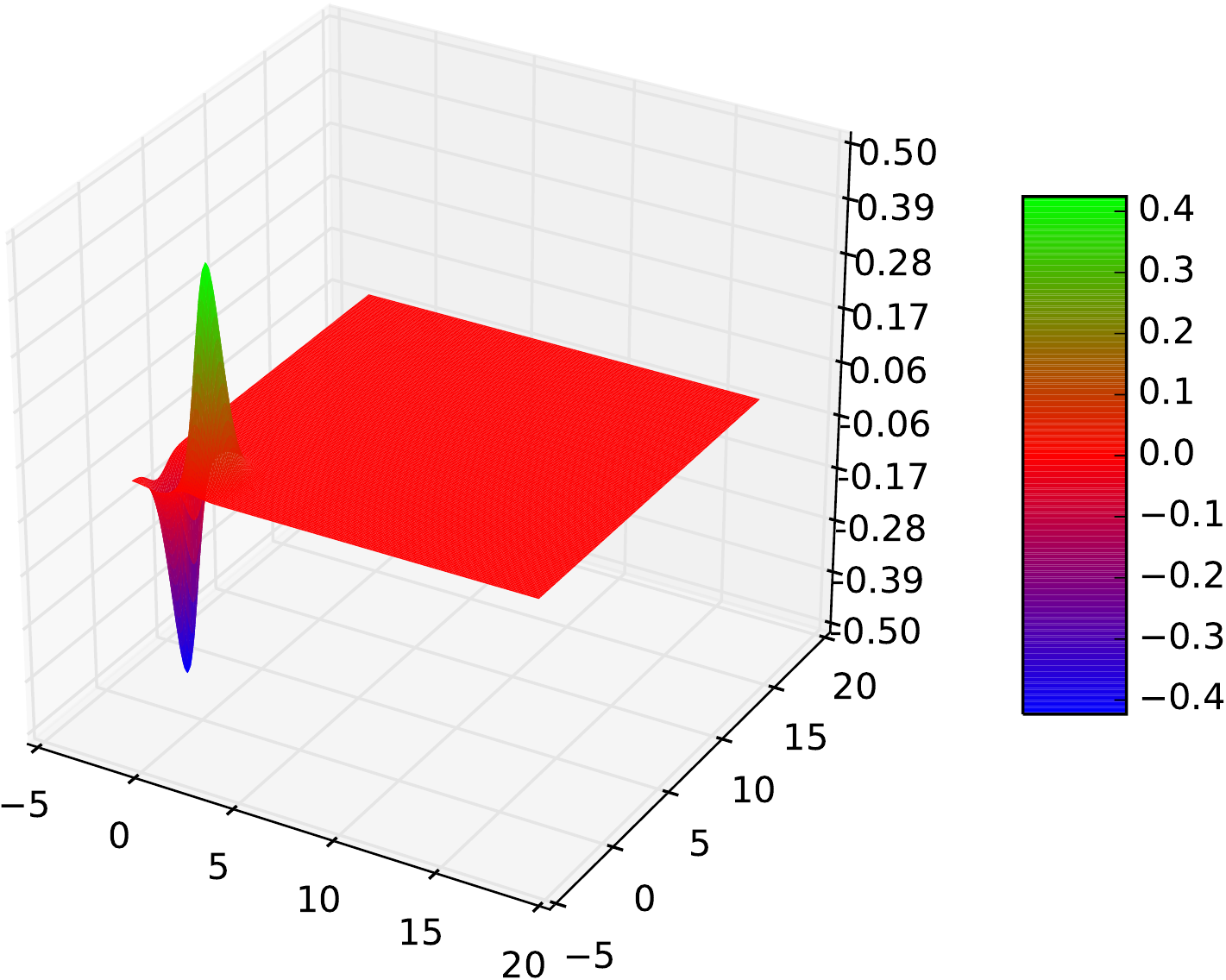}
  \includegraphics[width=0.35\linewidth]{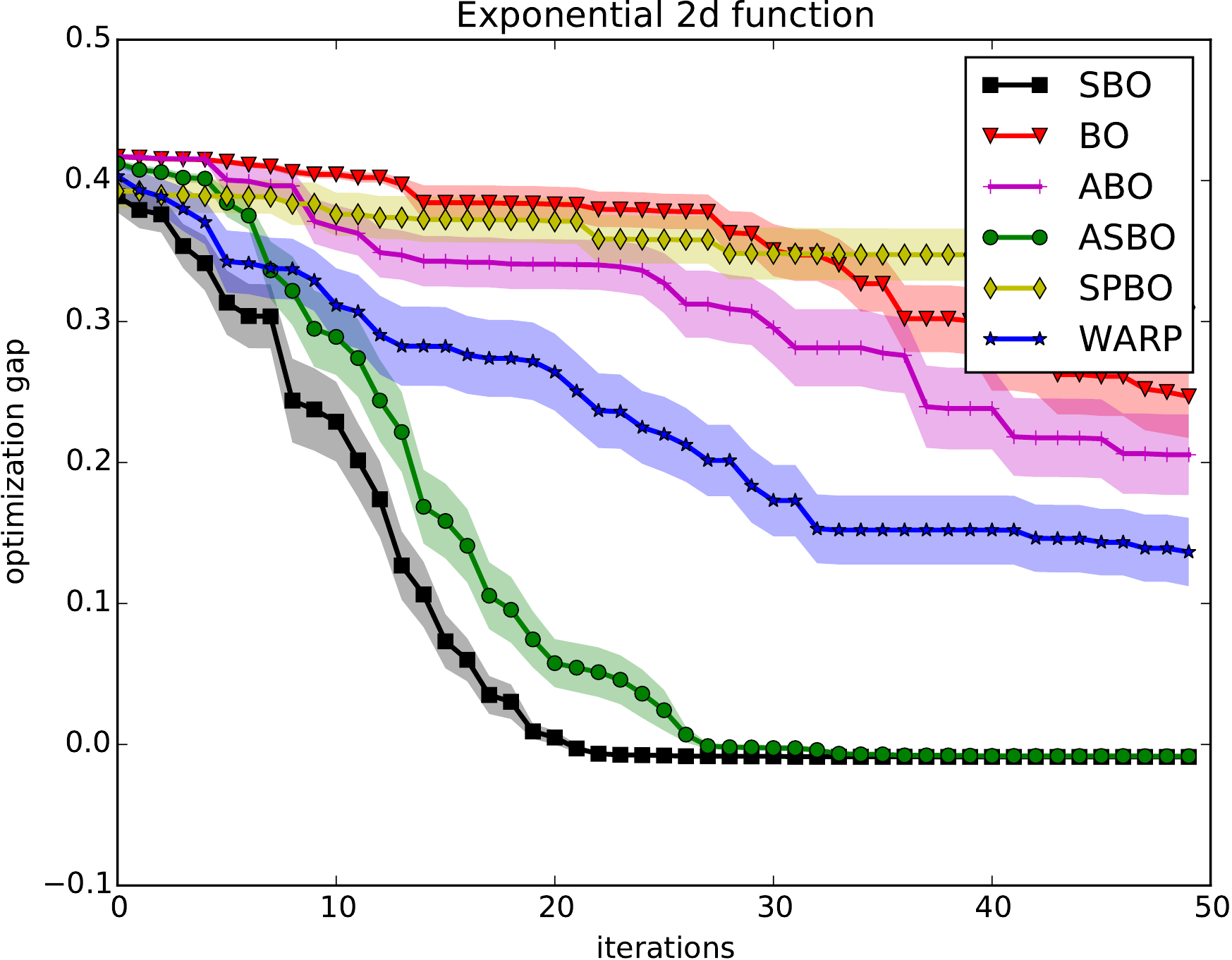}
  \includegraphics[width=0.35\linewidth]{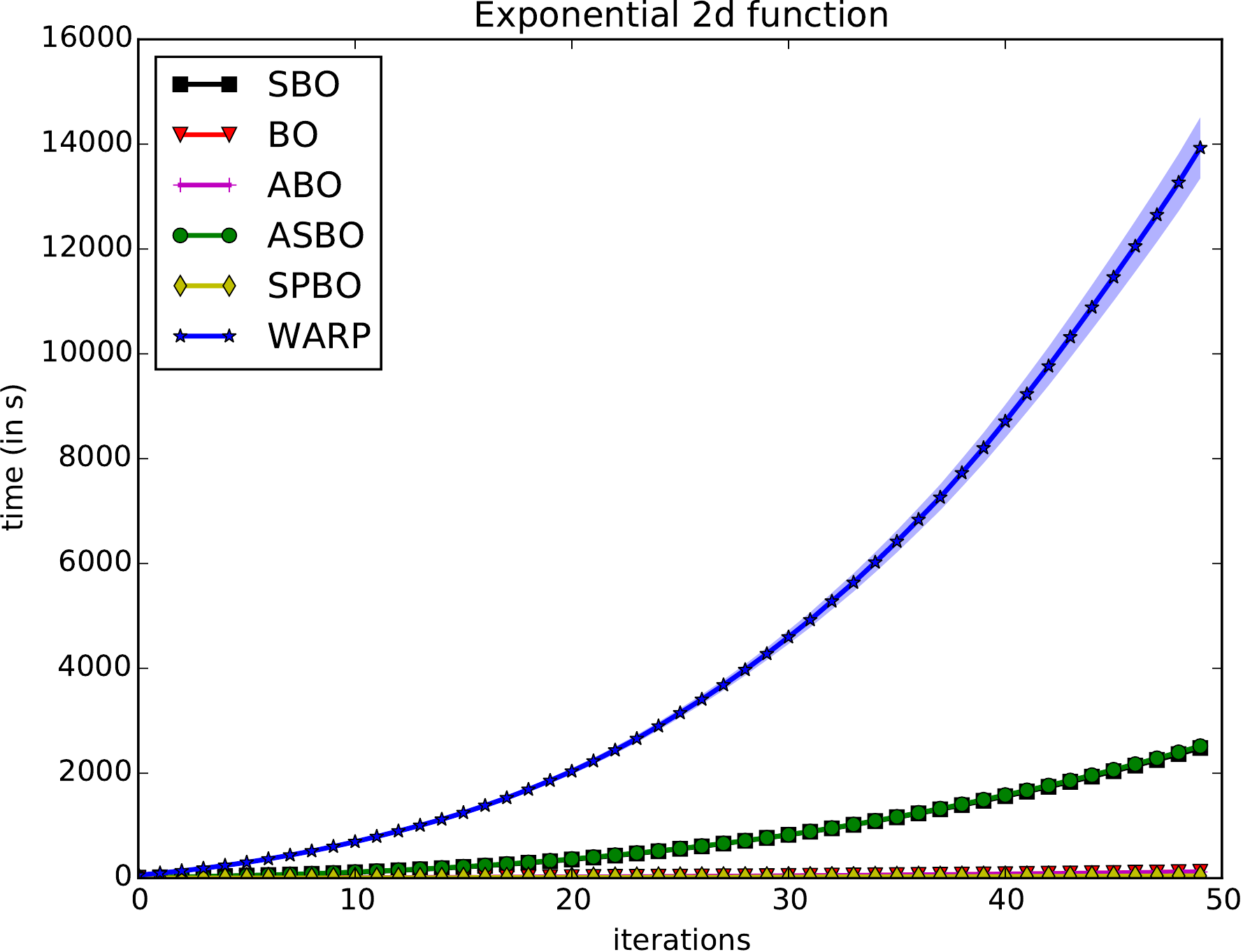}
  \caption{Results for the exponential 2D function. Center: we can see the optimization gap. Nonstationary methods, such as \emph{WARP} and the proposed method \emph{SBO} result in an improved convergence. The sugested method was able to find the minimum always in less than 20 iterations. Right: Total optimization time in seconds.}
  \label{fig:exp2dres}
\end{figure}

The intuition behind this setup is the same of many adquisition functions in Bayesian optimization: the aim of the surrogate model is not to approximate the target function precisely in every point, but to provide information about the location of the minimum. For example, the resulting model could ``flat out'' most of the search space, as soon as the region near the minimum have the correct variability. Many optimization problems are difficult due to the fact that the region near the minimum has higher variability that the rest of the space, like the function in Figure \ref{fig:exp2dres}. However, it is important to note that the kernel hyperparameters are initialized with the same prior for the local and global kernel. Thus, there is guarantee that the local kernel became the kernel with smaller length-scale. Depending on the data captured, it could learn a model where the local kernel has larger length-scale (i.e.: smoother) than the global kernel.

\section{Active hyperparameter learning during Bayesian optimization}

As commented previously, learning accurate kernel hyperparameters is of paramount importance in Bayesian optimization, specially in the case of nonstationary models, where the number of hyperparameters increases considerably. On the other hand, Bayesian optimization algorithms are rooted on the idea of having few samples. Therefore, learning those hyperparameters becomes a challenging problem.

In this section, we present a method to actively learn the kernel hyperparameters while doing optimization, by explicitly exploring areas with high information gain over the hyperparameters.

\subsection{Active hyperparameter learning through Information Gain}
 
In order to improve the quality of the surrogate model, we will try to reduce the entropy of the hyperparameters $H(\Theta)$. Implicitly, this is done already in Bayesian optimization. Many popular criteria such as the expected improvement, the upper confidence bound, etc., include an exploration term which tries to minimize the predicting variance of the surrogate model. The \emph{information never hurts} principle shows that any exploration strategy will, in expectation, decrease $H(\Theta)$. See Proposition 8 of \cite{krause07nonmyopic} for details.

Krause and Guestrin \cite{krause07nonmyopic} also suggest a criterion for explicit exploration based on the information gain (IG) of the GP hyperparameters. Following that criterion, the decision for the next point to evaluate will be:

\begin{equation}
  \label{eq:ige}
  \begin{split}
  c_{IG} &= H(\Theta | \y_{1:n}) - H(\Theta | \y_{1:n}, \y_{n+1}) = \arg \max_{\x_{n+1}} \; H(\y_{n+1} | \y_{1:n}) - H(\y_{n+1} | \y_{1:n}, \Theta) \\
          &\approx  - \log \left\{\sum_{i=1}^{m} \left[ \left(\mu_i - \widehat{\mu}\right)^2 + \sigma^2_i\right]\right\} + \sum_{i=1}^m \log \sigma^2_i
  \end{split}
\end{equation}
where $\widehat{\mu} = \sum_i \mu_i / m$. Any other sampling distribution could be used for $\Theta$ adding the corresponding weight to each sample. For example, in the original work \cite{krause07nonmyopic}, a discrete distribution is used.

The problem with $\av_{IG}$ is that it is purely exploratory criterion, which is not intended for optimization. Thus, it needs to be combined with other optimization criterion. We suggest a linear combination with an annealing coefficient to gather information about the hyperparameters $\Theta$ at the beginning, and focusing on finding the minimum $\x^*$ after several iterations. For example, the expected improvement with explicit information gain criterion is defined as:
\begin{equation}
  \label{eq:eiig}
  c_{EIIG}  = \sum_{i=1}^m \left[\left(\y_{best} - \mu_i\right) \Phi(z_i) + \sigma_i \phi(z_i)\right] + \frac{\alpha}{n^2}  \left(\log \left\{\sum_{i=1}^{m} \left[\left(\mu_i - \widehat{\mu}\right)^2 + \sigma^2_i\right]\right\} + \sum_{i=1}^m \log \sigma^2_i \right)
\end{equation}
where $z_i = (\y_{best} - \mu_i)/\sigma_i$. The functions $\Phi(\cdot)$ and $\phi(\cdot)$ are the normal cdf and pdf respectively. The coefficient $\alpha/n^2$ represents the importance   information gain component with respect to the expected improvement. Also, the coefficient is annealed to increase the effect of the IG early on to help learning the kernel hyperparameters. Eventually $\alpha/n^2 \rightarrow 0$, resulting in the classical expected improvement once the hyperparameters are good enough.

\subsection{Simultaneous perturbation Bayesian optimization}

Analyzing the shape of the information gain criterion we found that, in many cases, the highest value was near a previous sample. We concluded that the leght-scale hyperparameter is related to the variability of the function, which, at the same time, it is related to the local gradient. Therefore, in a certain way, the information gain criterion was estimating the gradient numerically to obtain information of the variability of the function. It is known that adding gradient information can improve considerably the accuracy of Gaussian process regression   \cite{Rasmussen:2006}. However, in many applications, the gradient is not available. Furthermore, using finite difference methods require many samples, which is against the philosophy of Bayesian optimization.

In the field of stochastic optimization, the simultaneous perturbation stochastic approximation algorithm (SPSA) \cite{Spall1998} is a variation of the classic finite difference algorithm for high dimensional spaces. In this method, the gradient is approximated by finite differences of a small subset of randomly sampled perturbations. The beauty of the SPSA algorithm is that, although the gradient is not perfectly estimated, the algorithm is guaranteed to converge to the local optimum. 

\begin{algorithm}
\caption{Simultaneous Perturbation Bayesian Optimization (SPBO)}\label{al:spbo}
\begin{algorithmic}[1]
\For{$i = 1 \ldots N/2$}  \Comment{Generates two samples per iteration}
   \State $\x_i \gets$  Computed with Algorithm \ref{al:bo}.
\If{$i < T$}\Comment{We compute perturbations only at the beginning, during exploration.}
   \State $\Delta_x$ perturbation vector. Each component sampled from a $\pm 1$ Bernoulli distribution 
   \State $\x_j \gets \x_i + \frac{c}{i^{\gamma}} \Delta_x$ 
\EndIf
\EndFor
\end{algorithmic}
\end{algorithm}

We present an algorithm called Simultaneous Perturbation Bayesian Optimization, which, for every sample using Bayesian optimization, also computes a perturbed sample. Theoretically, this algorithm reduce the computational burden of Bayesian optimization and information gain inasmuch as it is applied half of the iterations. Compared to Bayesian optimization, the cost of sampling a random perturbation is negligible with respect to the cost of updating the Gaussian process model and computing the maximum information gain. Also, note how the perturbation can be computed only in the first part of the optimization process, similar to the annealing process in the previous section.

\section{Hierarchical Bayesian optimization}
\label{sec:hbo}

In many Bayesian optimization applications, it is becoming a common trend to simultaneously optimize different kinds of variables, for example: continuous, discrete, categorical, etc. While Gaussian processes are suitable for modeling those spaces, Bayesian optimization can become quite involved in line \ref{al:crit} of Algorithm \ref{al:bo}, as the acquisition function must be optimized in the same input space. Available implementations of Bayesian optimization like Spearmint \cite{Snoek2012} use grid sampling and rounding tricks to combine different input spaces. However, this might reduce the quality of the final result \cite{MartinezCantin14jmlr} compared to proper optimization methods.

In this section, we suggest to use a hierarchical Bayesian optimization model, where the input space is partitioned between homogeneous variables, for example: continuous variables $\x^{(c)}$ and discrete variables $\x^{(d)}$. Therefore, the evaluation of an element higher in the hierarchy implies the full optimization of the elements lower in the hierarchy. In principle, that would require much more function evaluations but, as the input space has been partitioned, the dimensionality of each separate problem is much lower. In practice, for the same number of function evaluations, the computational cost is considerably reduced.

\begin{algorithm}
\caption{Hierarchical Bayesian Optimization (HBO) |  Total budget $N = N_1 \cdot N_2$}\label{al:hbo}
\begin{algorithmic}[1]
\State $\x = [\x^{(c)},  \x^{(d)}]$ \Comment{We separate the discrete and continuous components}
\For{$n = 1 \ldots N_c$} \Comment{Outer loop - Continuous optimization}
   \State Update model and kernels hyperparameters for $\x^{(c)}$
   \State Find continuous component of next point $\x^{(c)}_{n}$
   \For{$k = 1 \ldots N_d$} \Comment{Inner loop - Discrete optimization}
         \State Update model and kernels hyperparameters for $\x^{(d)}$
         \State Find discrete component of next point $\x^{(d)}_{k}$
         \State $\y_k \gets f([\x^{(c)}_{n}, \x^{(d)}_{k}])$ \Comment{Update $\x^{*(d)}_n$}
   \EndFor
   \State $\y_n \gets f([\x^{(c)}_{n}, \x^{*(d)}_n])$ \Comment{Update $\x^{*(c)}$ and save corresponding $\x^{*(d)}$}
\EndFor
\State $\x^* = [\x^{*(c)},  \x^{*(d)}]$
\end{algorithmic}
\end{algorithm}

An advantage of this approach is that we can combine different algorithms for different levels of the hierarchy. For example, using Random Forests \cite{HutHooLey11-smac} could be more suitable as a surrogate model for certain discrete/categorical variables than Gaussian processes. In contrast, we lose the correlation among variables in the inner loop, which might be counterproductive in certain situations.

\section{Evaluation and results}
\label{sec:results}

\begin{figure}
  \centering
  \includegraphics[width=0.31\linewidth]{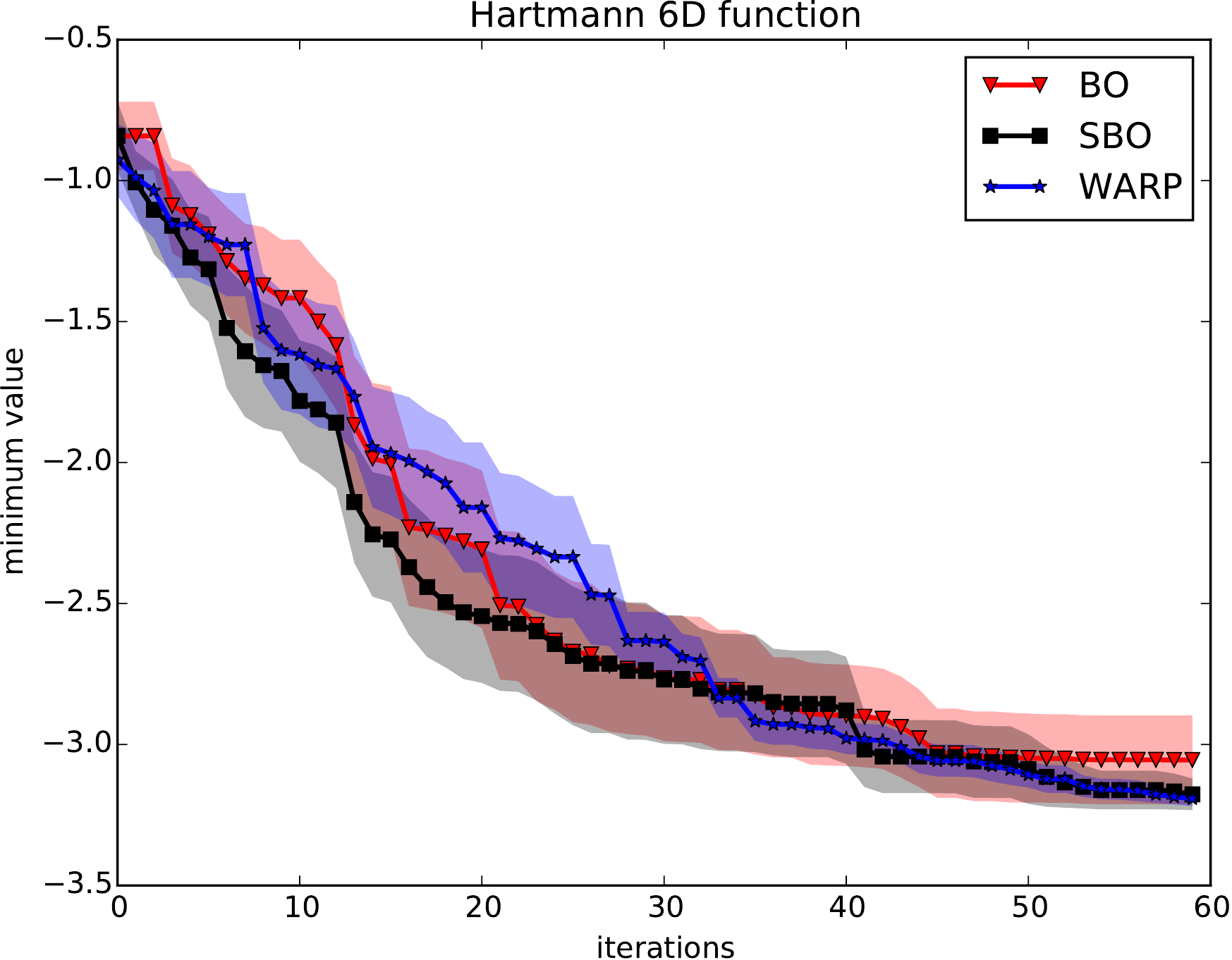}
  \includegraphics[width=0.31\linewidth]{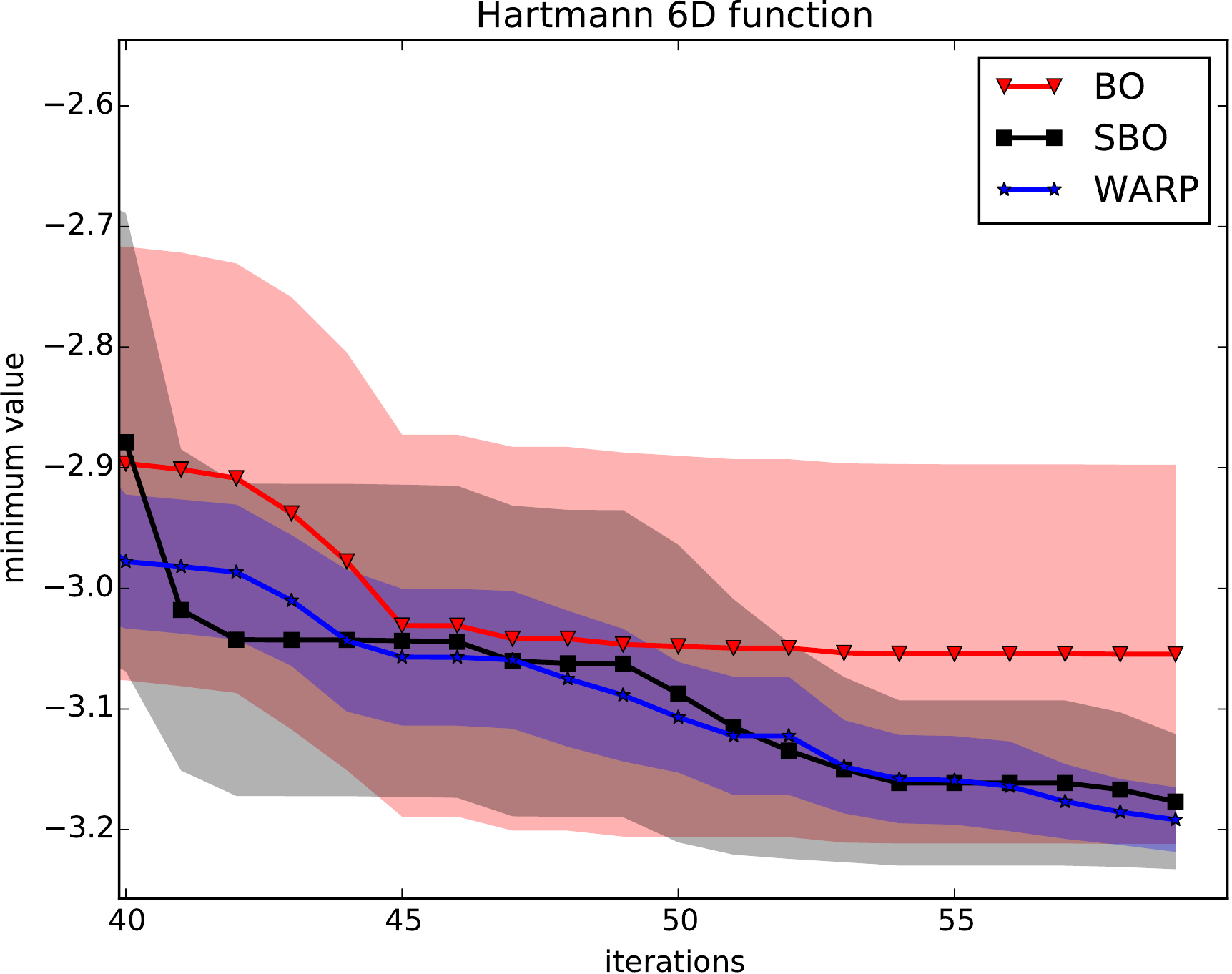}
  \includegraphics[width=0.31\linewidth]{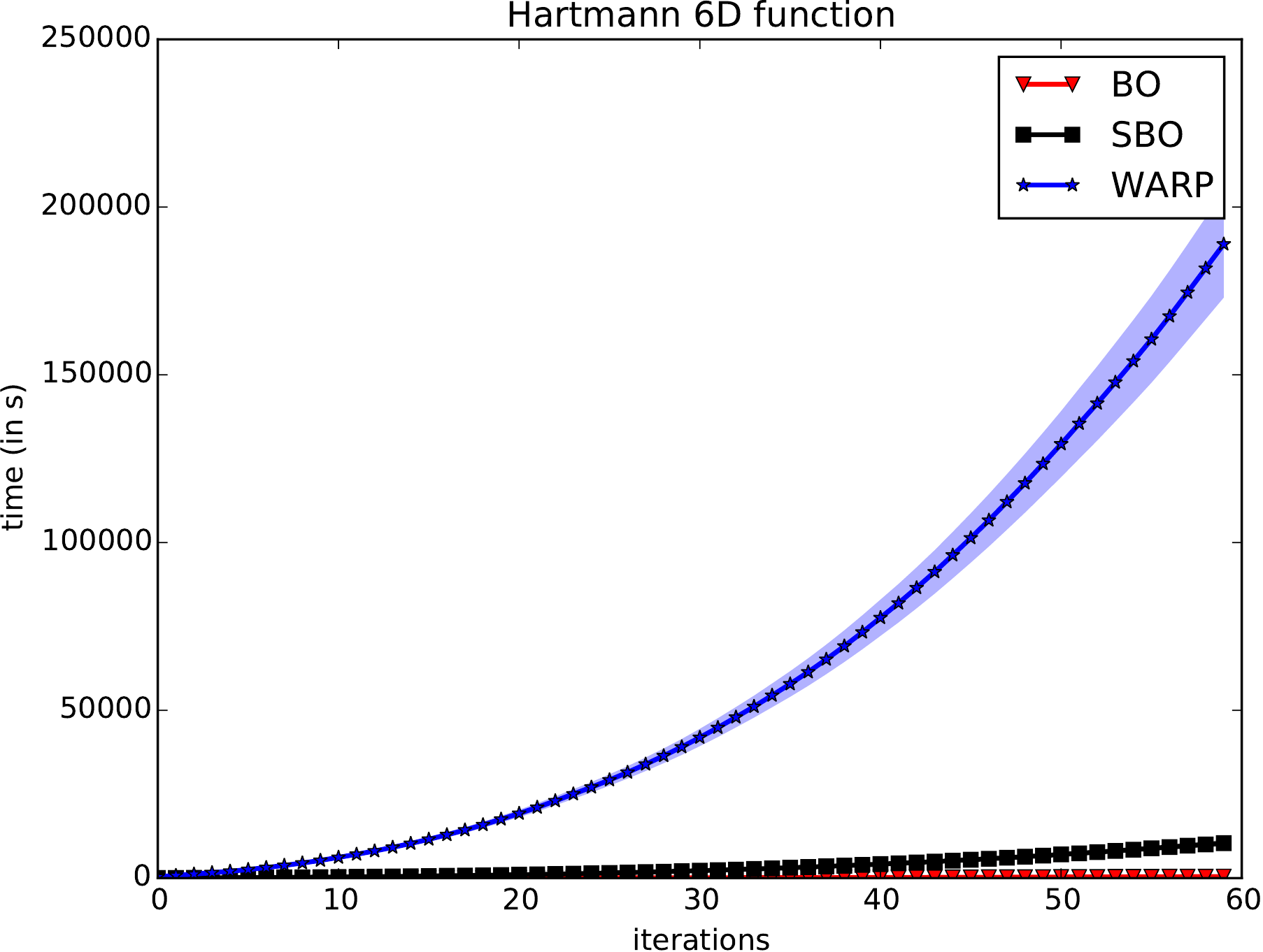}
  \caption{Hartmann 6D function. Left: Minimum value obtained. Center: zoom of the minimum value in the last iterations. Right: time in seconds. }
  \label{fig:hartmann}
\end{figure}

For the evaluation of the suggested algorithms we have implemented them using the popular BayesOpt\footnote{The code will be available as GPL once the paper gets published.} library \cite{MartinezCantin14jmlr}. This allowed us to compare our proposal with a large variety of surrogate models, kernels, etc. For example, the results presented in this sections are based on the standard convention in Bayesian optimization literature, that is, a simple zero-mean Gaussian process, a Mat{\'e}rn kernel 5/2 with automatic relevant determination for continuous variables, a Hamming kernel as presented in \cite{ZiyuWang2013} for categorical variables and slice sampling for learning the model hyperparameters (kernel, warping, etc.). However, the suggested method has also been tested with other models such as Student-t processes, other kernels, etc. Due to the computational burden of MCMC for the hyperparameters, we have used a small number of samples (10), while trying to decorrelate every resample by large burn-in periods (100 samples) following the convention in \cite{Snoek2012}.

We compare standard Bayesian Optimization (BO), Bayesian Optimization with Active hyperparameter learning (ABO), Spartan Bayesian Optimization (SBO), Spartan Bayesian Optimization with Active hyperparameter learning (ASBO), Simultaneous Perturbation Bayesian Optimization (SPBO). We also implemented the input warping (WARP) of Snoek et al. \cite{snoek-etal-2014a} which, to the authors knowledge, is currently the only Bayesian optimization algorithm specific for nonstationary functions. All experiments were repeated between 10 and 25 times, depending on the problem, using common random number to reduce the sampling error between algorithms. All the experiments include 5-10 initial samples generated through latin hypercube sampling, not included in the plot.

\subsection{Optimization benchmarks}

\begin{figure}
  \centering
  \includegraphics[width=0.31\linewidth]{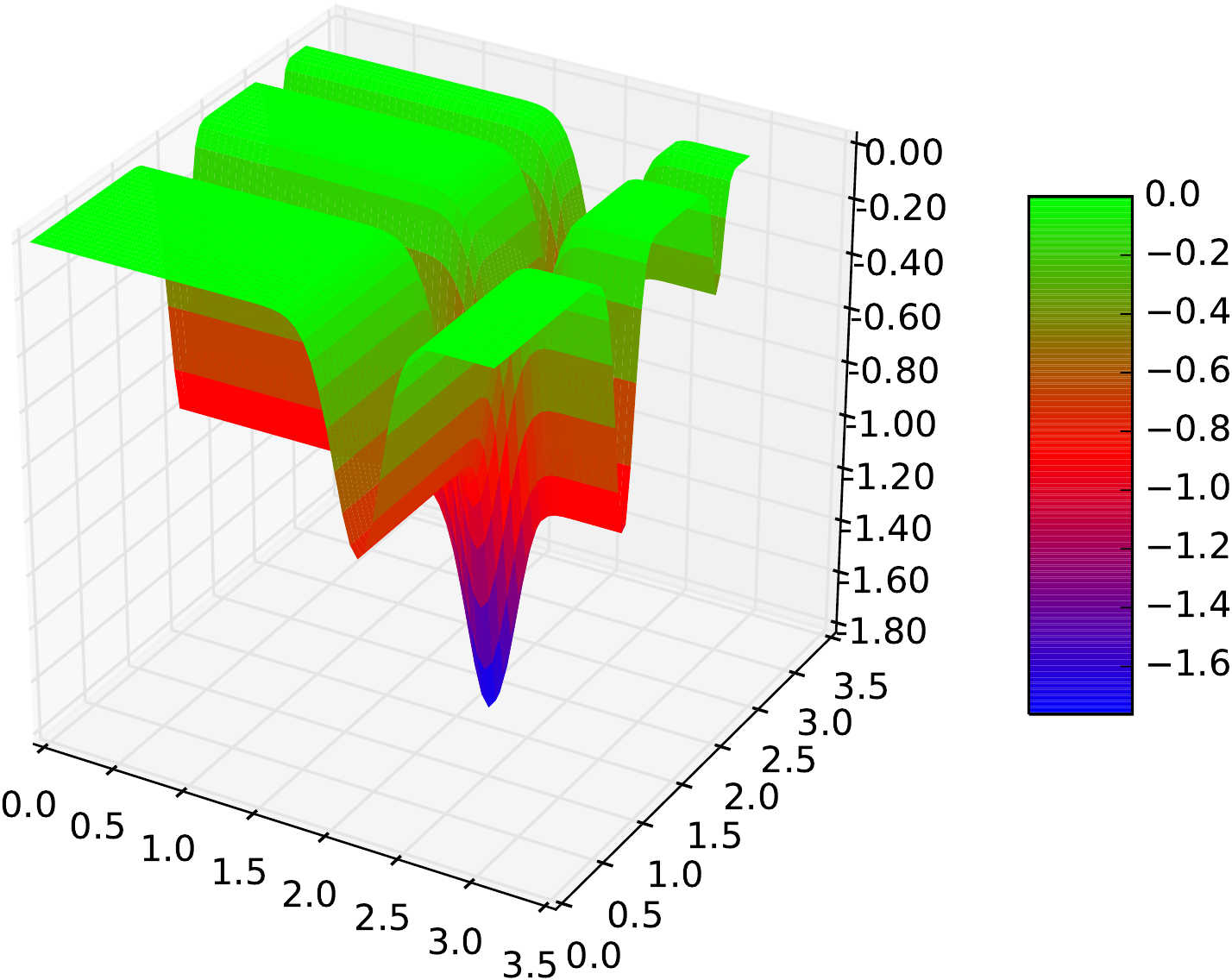}
  \includegraphics[width=0.31\linewidth]{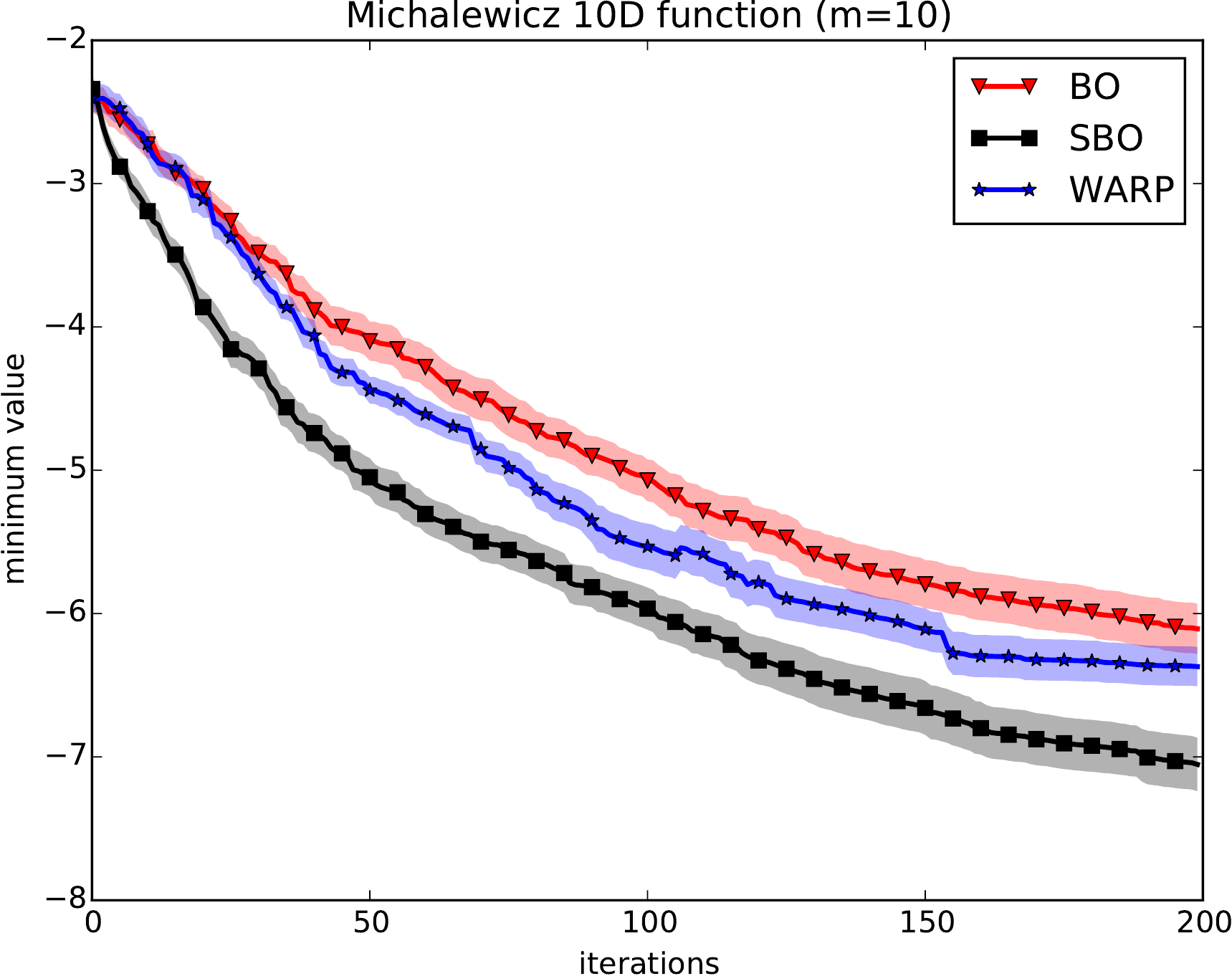}
  \includegraphics[width=0.31\linewidth]{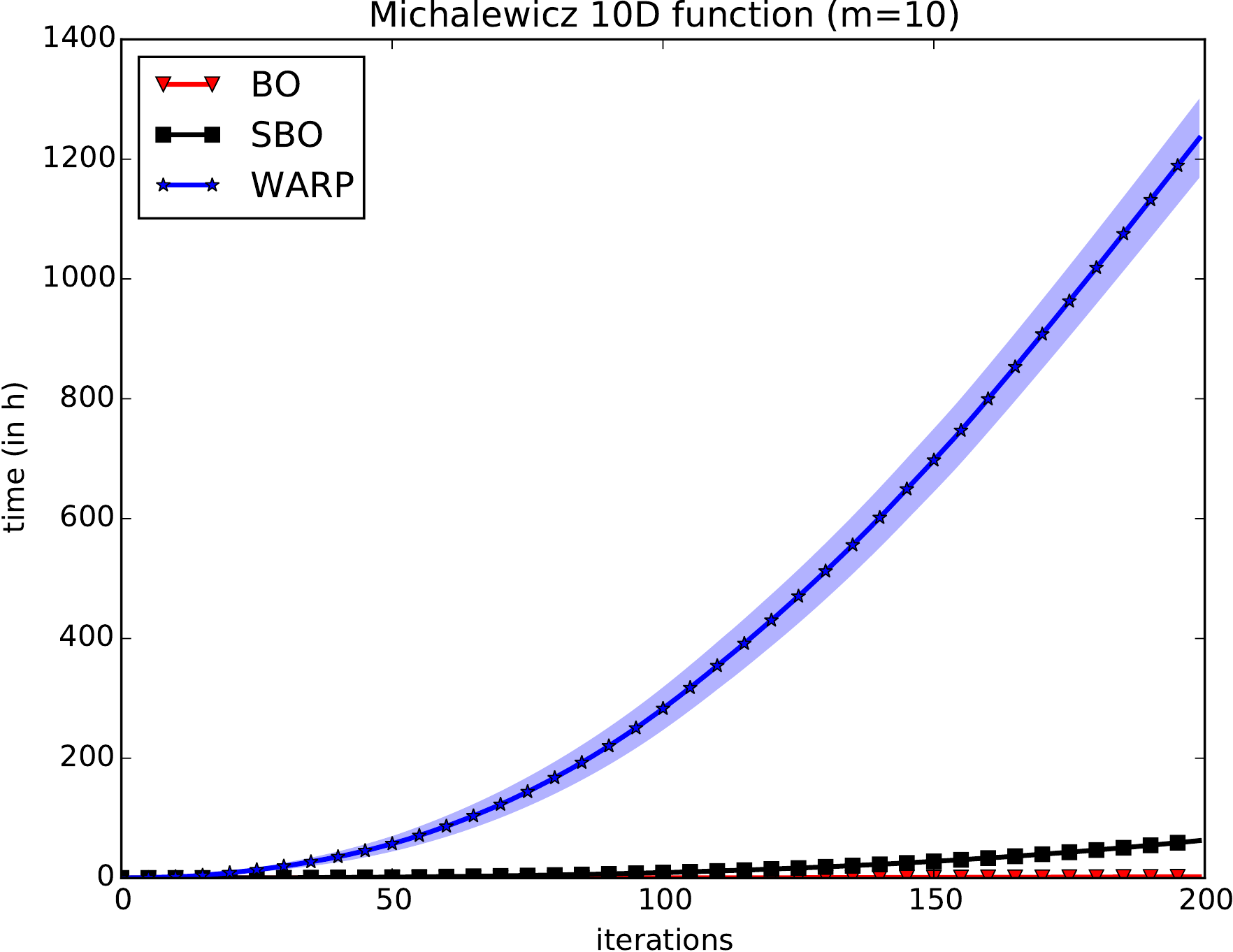}
  \caption{Michalewicz 10D function with m=10. On the left, there is a visualization of the first and second dimensions to appreciate the difficulty of this function. The number of local minima increases exponentially with the number of dimensions. Note that the time is in hours. }
  \label{fig:micha}
\end{figure}

We have evaluated the algorithms on a set of standard functions for global optimization. Although popular in the literature, we have avoided the use of the Branin-Hoo and Camelback functions as there is barely room from improvement using ``vanilla'' BayesOpt \cite{MartinezCantin14jmlr}.

Figure \ref{fig:exp2dres} shows the results of optimizing the exponential 2D function $f(\x) = x_1 \exp(-x_1^2-x_2²)$ for ${x_1, x_2} \in [-2,18]^2$ from \cite{gramacy2005bayesian}. The use of classical stationary models (BO) results in a poor convergence because of the high nonstationarity of the function. Using an active strategy speed up the results, requiring slightly fewer iterations. Clearly, nonstationary methods, such as Snoek et al \cite{snoek-etal-2014a} (WARP) and the proposed method (SBO) result in an improved convergence. In fact, the convergence of the proposed method requires a number of iterations so small that actively learning the kernel hyperparameters results in a waste of samples.

The computation times of the different algorithms are clearly driven by the dimensionality and shape of the distribution of the MCMC. The cost of the active component is negligible when compared to their passive counterpart. We found that learning the Beta parameters for the warping function was extremely slow as many samples were rejected. This could be alleviated with a different MCMC algorithm such as hybrid Monte Carlo or sequential Monte Carlo samplers. However, note that we use slice sampling as recommended by the authors \cite{snoek-etal-2014a}.

For the Hardmann 6D function (shown in Figure \ref{fig:hartmann}), the differences are not statistically significant, which shows that when the function is stationary, nonstationary methods are no worse than standard Bayesian optimization. Furthermore, we can see that at the end they are more robusts.

The Michalewicz function is known to be one of the hardest benchmarks in surrogate based global optimization. Figure \ref{fig:micha} shows the results. The active algorithms as well as the SPBO have been removed as they were statistically indifferent from their passive counterpart.

\subsection{Surrogate benchmarks for machine learning problems}

\begin{figure}
  \centering
  \includegraphics[width=0.32\linewidth]{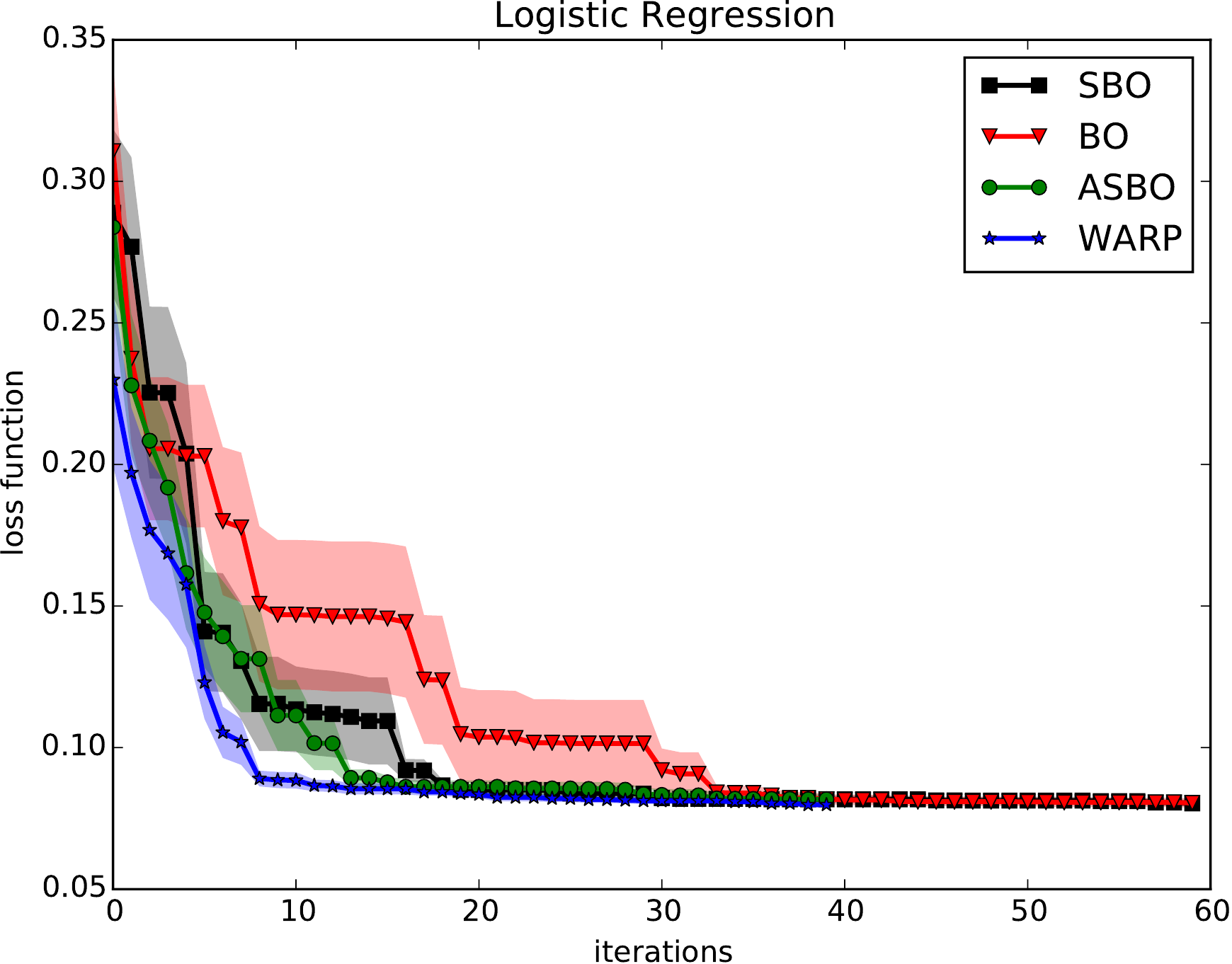}
  \includegraphics[width=0.32\linewidth]{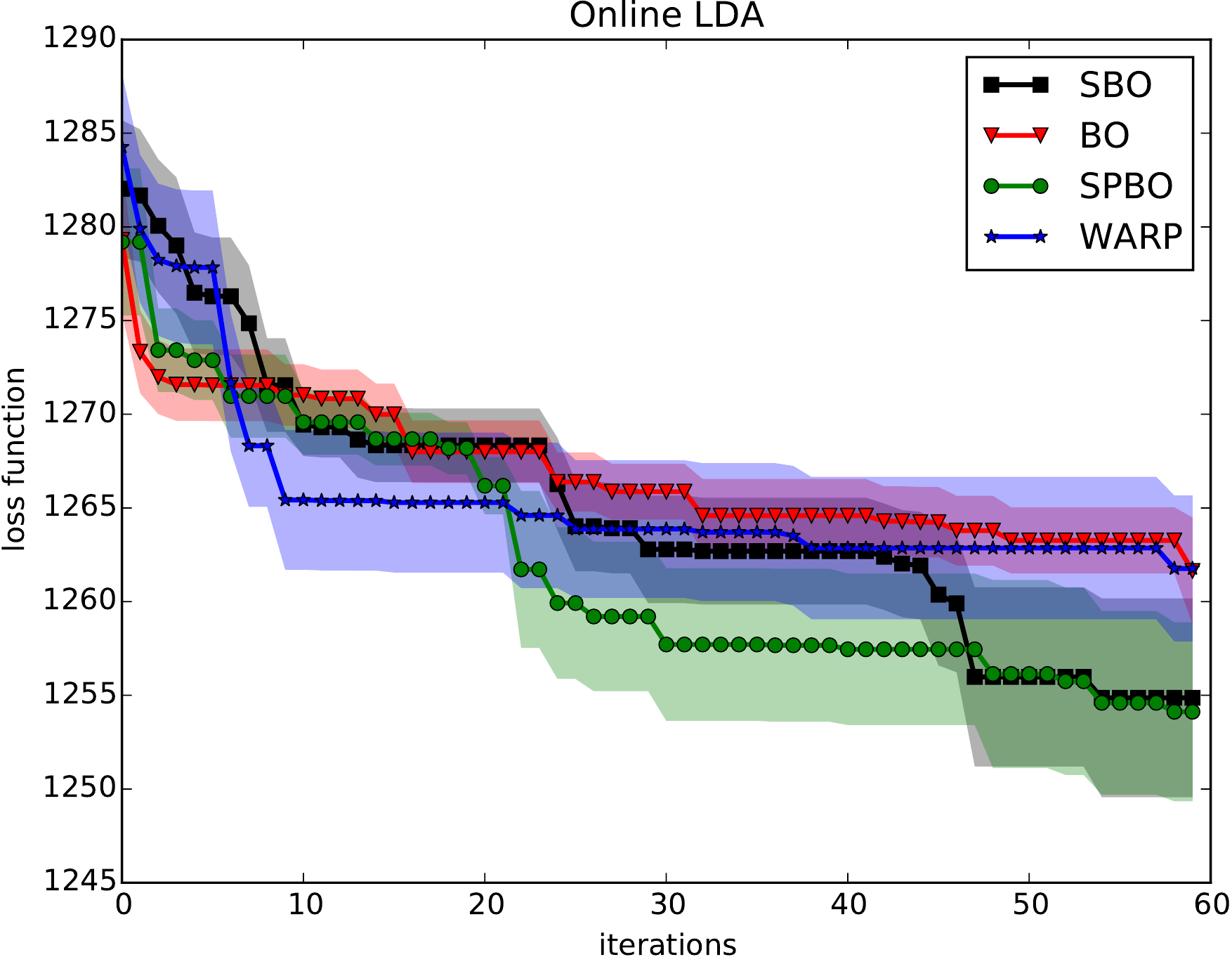}
  \includegraphics[width=0.32\linewidth]{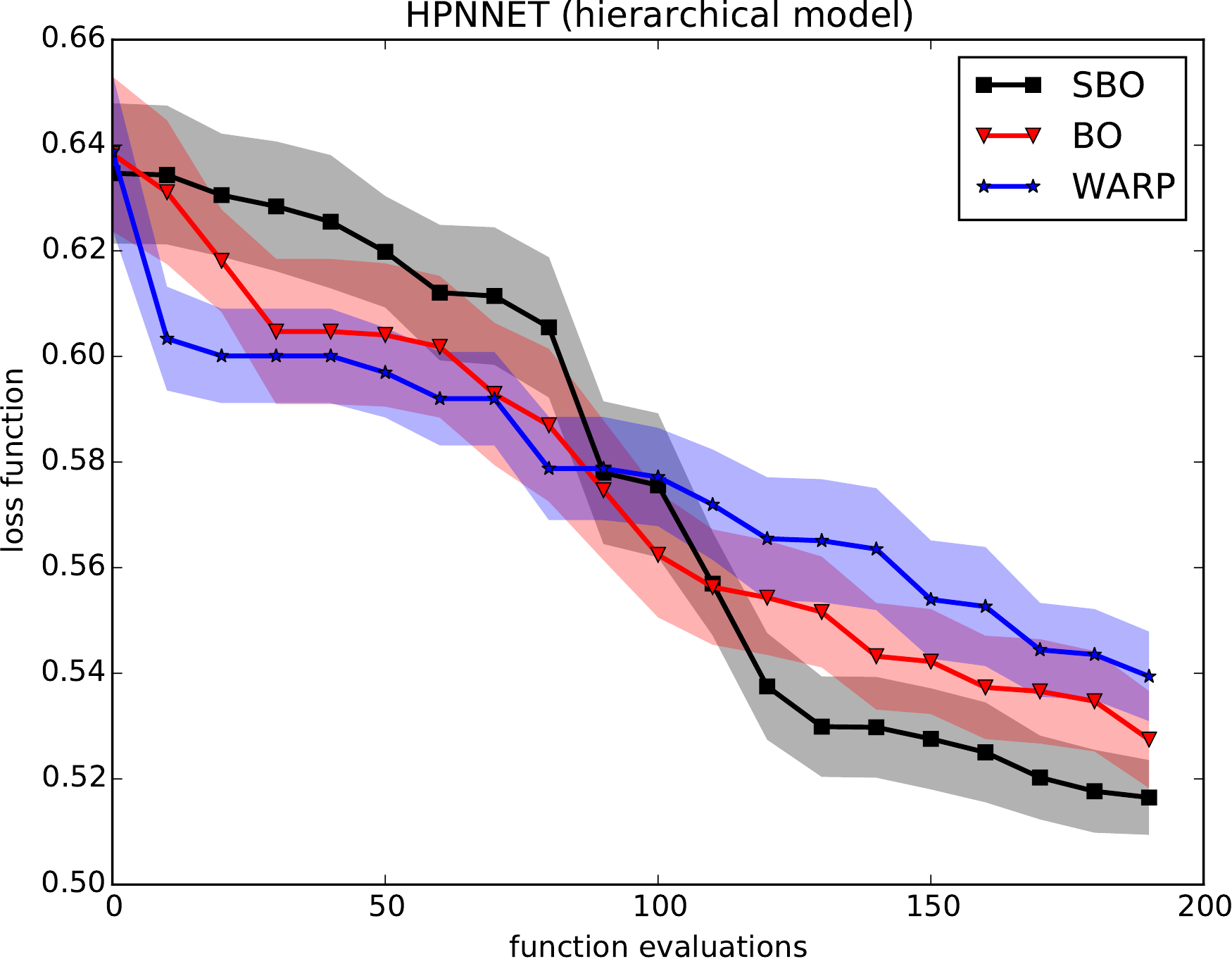}
  \\ \vspace{3pt}
  \includegraphics[width=0.32\linewidth]{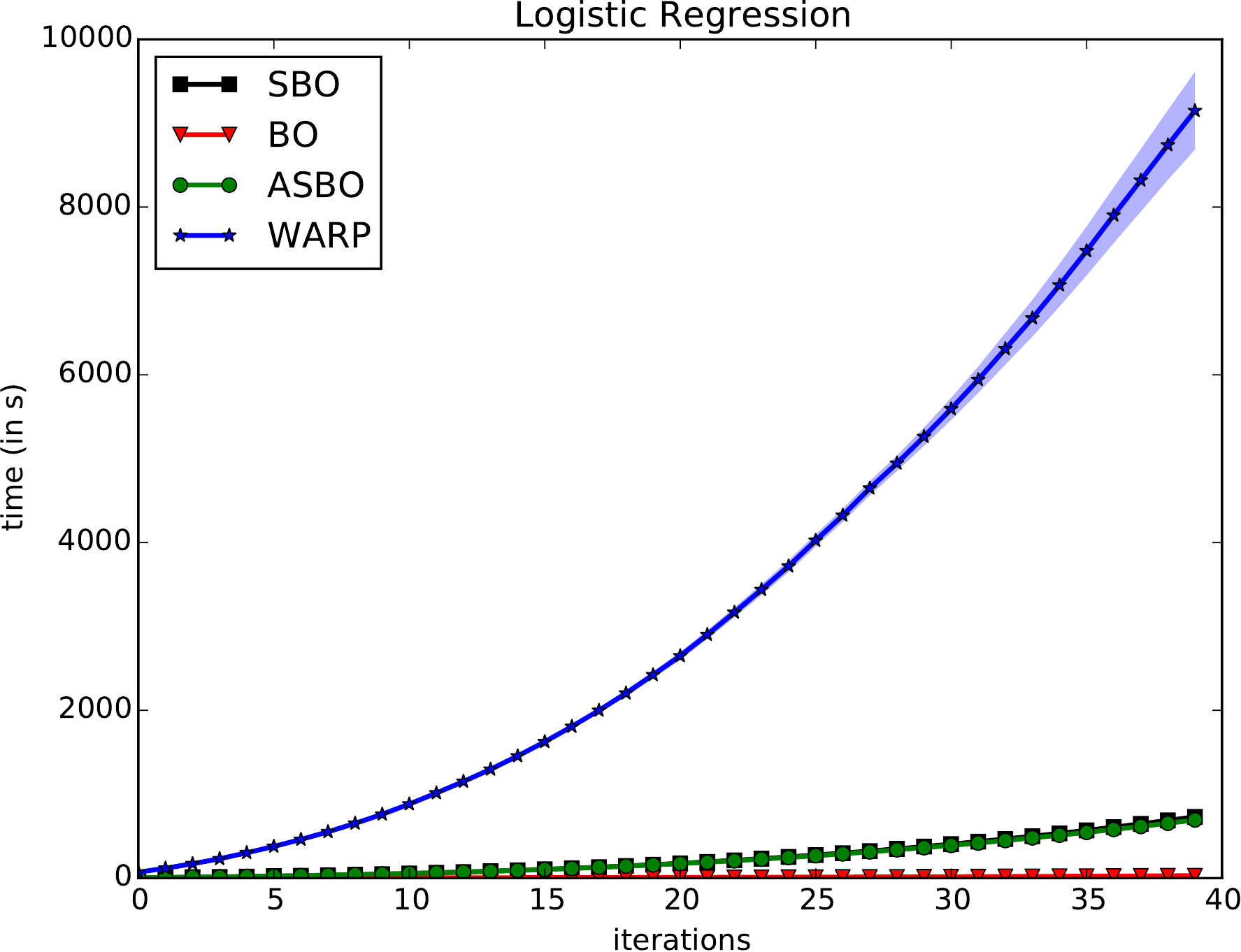}
  \includegraphics[width=0.32\linewidth]{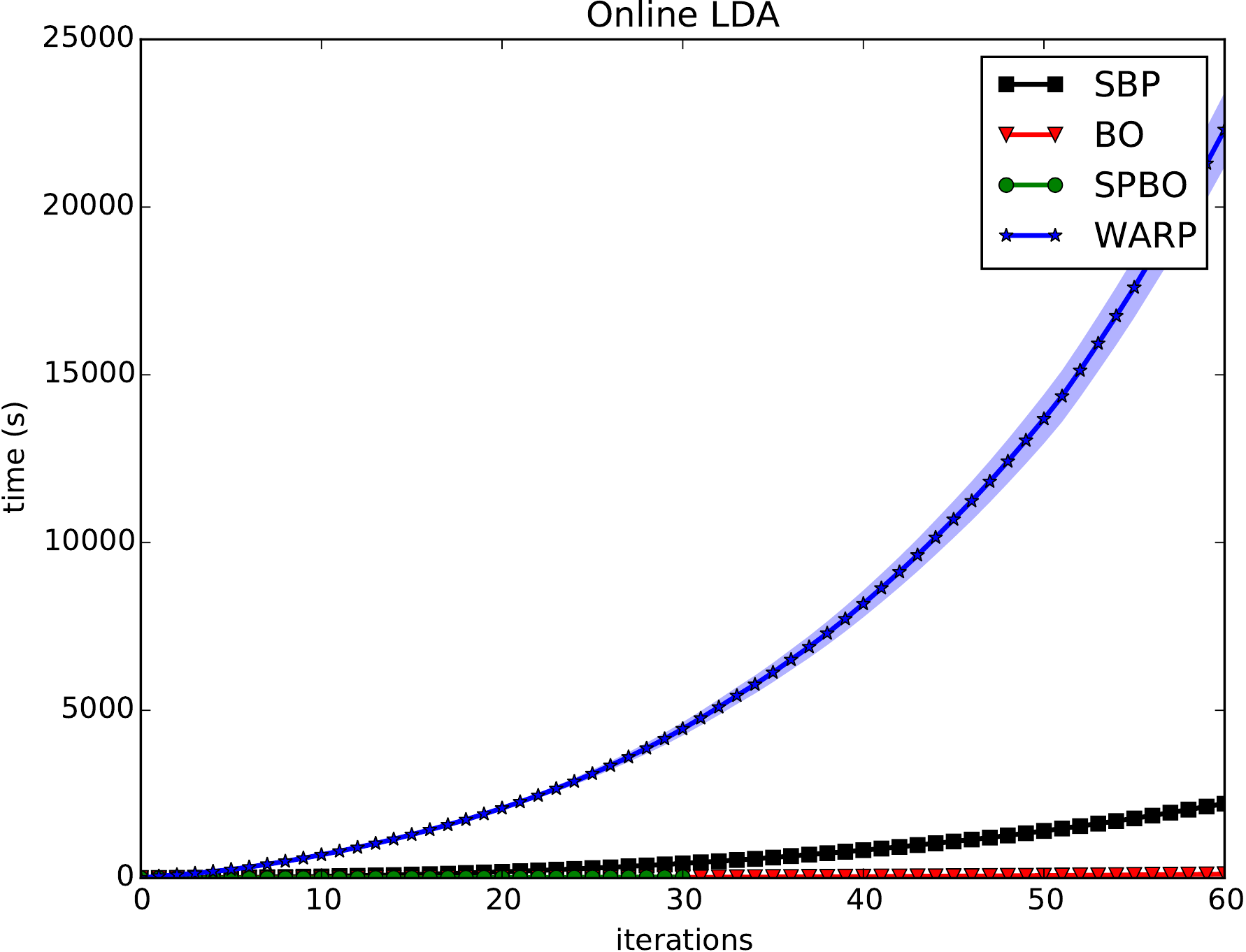}
  \includegraphics[width=0.32\linewidth]{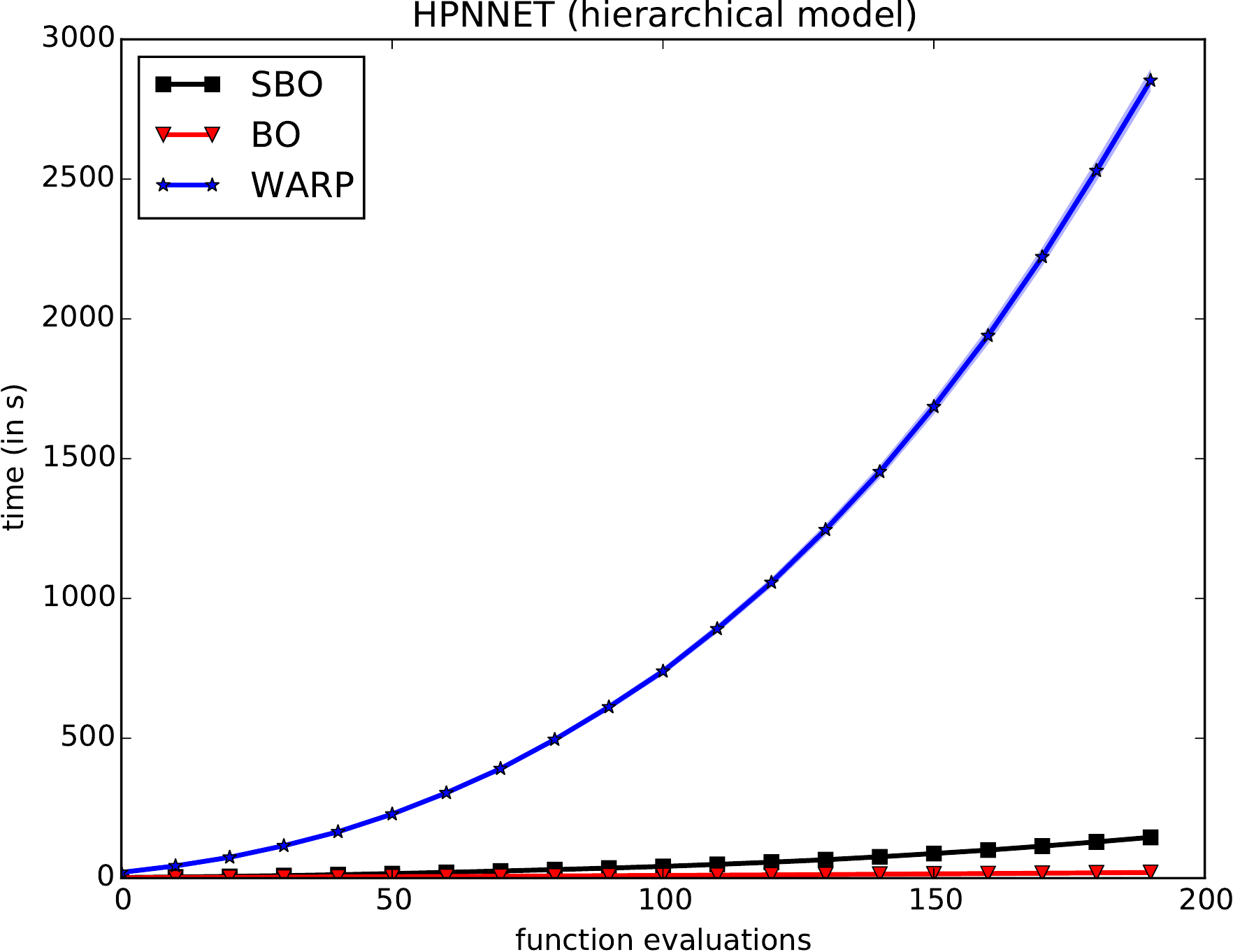}
  \caption{Surrogate benchmarks  \cite{Eggensperger2015} based on real hyperparameter optimization of machine learning algorithms. From left to right: logistic regression (4D continuous), online LDA (3D continuous) and deep neural network (HP-NNET with the mrbi dataset, 7D continuous, 7D categorical). Top row: loss functions. Bottom row: computational time, in seconds.}
  \label{fig:results}
\end{figure}

Our next set of experiments is based on well known benchmarks for automatic tunning of machine learning problems. However, in order to simplify the evaluation, we have used the surrogate benchmarks presented in \cite{Eggensperger2015}. Among all the available benchmarks we have selected the Gradient Boosting as it provides the lowest RMSE with respect to the actual problem. We explicitly avoid the Gaussian process surrogate to avoid the advantage of perfectly modeling. The results are shown in Figure \ref{fig:results}. For clarity, we show only the statistically significant results apart from BO, local and warp.

First, the logistic regression is an easy problem for Bayesian optimization. Even the simple Bayesian optimization can reach the minimum in less than 40 iterations. In this case, the warped method is the fastest one, with less than 10 iterations. However, the proposed method is comparable specially when using active learning, by a fraction of the total cost. It is important to note that, although Bayesian optimization is intended for expensive functions and the cost per iteration is negligible, we are talking of minutes to hours for single iteration of the warped method in a standard laptop. For the onlineLDA problem, the warped method gets stuck like the standard Bayesian optimization, while our method is able to escape from the local minima. Surprisingly, the SPBO method is the best in this example, even though the computational cost is the lowest by a large margin. This shows that for certain applications and structured problems, gradient information is fundamental.

Finally, we evaluate the HP-NNET problem. In this case, due to the high dimensionality and heterogeneity of the input space (7 continuous + 7 categorical parameters) we have applied the hierarchical model presented in Section \ref{sec:hbo}. Also, in order to reduce the computational cost, the nonstationary (warping or local kernel) is applied only on the continuous variables (outer loop). Note that, in this case, the plots are with respect to target function evaluations. In this case, due to the complexity of the problem, the local kernel fails to converge at an early stage. However, with more data available, the local kernel jumps to a good spot and the convergence is faster.

\section{Conclusions}

We have presented a new algorithm called Spartan Bayesian Optimization which combines a local and a global kernel to deal with nonstationarity in Bayesian optimization. Besides, we have shown that the model can improve convergence speed even in stationary problems. The suggested algorithm achieves similar or better results than the state of the art by a small fraction of the computational cost. In addition to that, we have presented some ideas to improve the results of any Bayesian optimization algorithm: by actively learning the surrogate hyperparameters, by efficiently estimating the gradient of the target function and by building a hierarchical model to reduce the heterogeneity in the input space.




\small{
\bibliographystyle{plain}
\bibliography{../bib/optimization}

\begin{thebibliography}{10}

\bibitem{Borji2013}
Ali Borji and Laurent Itti.
\newblock Bayesian optimization explains human active search.
\newblock In C.J.C. Burges, L.~Bottou, M.~Welling, Z.~Ghahramani, and K.Q.
  Weinberger, editors, {\em Advances in Neural Information Processing Systems
  26}, pages 55--63. Curran Associates, Inc., 2013.

\bibitem{Bull2011}
Adam~D. Bull.
\newblock Convergence rates of efficient global optimization algorithms.
\newblock {\em Journal of Machine Learning Research}, 12:2879--2904, 2011.

\bibitem{Cully2015}
Antoine Cully, Jeff Clune, Danesh Tarapore, and Jean-Baptiste Mouret.
\newblock Robots that can adapt like animals.
\newblock {\em Nature}, 521:503–507, 2015.

\bibitem{Eggensperger2015}
K.~Eggensperger, F.~Hutter, H.H. Hoos, and K.~Leyton-Brown.
\newblock Efficient benchmarking of hyperparameter optimizers via surrogates.
\newblock In {\em Proceedings of the Twenty-Ninth AAAI Conference on Artificial
  Intelligence}, January 2015.

\bibitem{gramacy2005bayesian}
Robert~B Gramacy.
\newblock {\em Bayesian treed Gaussian process models}.
\newblock PhD thesis, University of California, Santa Clara, 2005.

\bibitem{HennigSchuler2012}
Philipp Hennig and Christian~J. Schuler.
\newblock Entropy search for information-efficient global optimization.
\newblock {\em Journal of Machine Learning Research}, 13:1809--1837, 2012.

\bibitem{NIPS2014_5324}
Jos\'{e}~Miguel Hern\'{a}ndez-Lobato, Matthew~W Hoffman, and Zoubin Ghahramani.
\newblock Predictive entropy search for efficient global optimization of
  black-box functions.
\newblock In Z.~Ghahramani, M.~Welling, C.~Cortes, N.D. Lawrence, and K.Q.
  Weinberger, editors, {\em Advances in Neural Information Processing Systems
  27}, pages 918--926. Curran Associates, Inc., 2014.

\bibitem{HutHooLey11-smac}
Frank Hutter, Holger~H. Hoos, and Kevin Leyton-Brown.
\newblock Sequential model-based optimization for general algorithm
  configuration.
\newblock In {\em LION-5}, page 507–523, 2011.

\bibitem{Jones:1998}
Donald~R. Jones, Matthias Schonlau, and William~J. Welch.
\newblock Efficient global optimization of expensive black-box functions.
\newblock {\em Journal of Global Optimization}, 13(4):455--492, 1998.

\bibitem{krause07nonmyopic}
Andreas Krause and Carlos Guestrin.
\newblock Nonmyopic active learning of gaussian processes: an
  exploration-exploitation approach.
\newblock In {\em International Conference on Machine Learning (ICML)},
  Corvallis, Oregon, June 2007.

\bibitem{MartinezCantin14jmlr}
Ruben Martinez-Cantin.
\newblock {BayesOpt}: A {Bayesian} optimization library for nonlinear
  optimization, experimental design and bandits.
\newblock {\em Journal of Machine Learning Research}, 15:3735--3739, 2014.

\bibitem{MartinezCantin09AR}
Ruben Martinez-Cantin, Nando {de Freitas}, Eric Brochu, Jose Castellanos, and
  Arnoud Doucet.
\newblock A {Bayesian} exploration-exploitation approach for optimal online
  sensing and planning with a visually guided mobile robot.
\newblock {\em Autonomous Robots}, 27(3):93--103, 2009.

\bibitem{Mockus1989}
Jonas Mockus.
\newblock {\em Bayesian Approach to Global Optimization}, volume~37 of {\em
  Mathematics and Its Applications}.
\newblock Kluwer Academic Publishers, 1989.

\bibitem{O'Hagan1992}
Anthony O'Hagan.
\newblock Some {B}ayesian numerical analysis.
\newblock {\em Bayesian Statistics}, 4:345--363, 1992.

\bibitem{Rasmussen:2006}
Carl~E. Rasmussen and Christopher~K.I. Williams.
\newblock {\em {G}aussian Processes for Machine Learning}.
\newblock The {MIT} Press, Cambridge, Massachusetts, 2006.

\bibitem{sampson1992nonparametric}
Paul~D Sampson and Peter Guttorp.
\newblock Nonparametric estimation of nonstationary spatial covariance
  structure.
\newblock {\em Journal of the American Statistical Association},
  87(417):108--119, 1992.

\bibitem{AmarShah2014}
Amar Shah, Andrew~Gordon Wilson, and Zoubin Ghahramani.
\newblock {Student-t} processes as alternatives to {G}aussian processes.
\newblock In {\em AISTATS, JMLR Proceedings. JMLR.org}, 2014.

\bibitem{Snoek2012}
Jasper Snoek, Hugo Larochelle, and Ryan Adams.
\newblock Practical {B}ayesian optimization of machine learning algorithms.
\newblock In {\em NIPS}, pages 2960--2968, 2012.

\bibitem{snoek-etal-2014a}
Jasper Snoek, Kevin Swersky, Richard~S. Zemel, and Ryan~Prescott Adams.
\newblock Input warping for bayesian optimization of non-stationary functions.
\newblock In {\em International Conference on Machine Learning}, 2014.

\bibitem{Spall1998}
J.C. Spall.
\newblock Implementation of the simultaneous perturbation algorithm for
  stochastic optimization.
\newblock {\em Aerospace and Electronic Systems, IEEE Transactions on},
  34(3):817--823, Jul 1998.

\bibitem{Srinivas10}
Niranjan Srinivas, Andreas Krause, Sham Kakade, and Matthias Seeger.
\newblock {G}aussian process optimization in the bandit setting: No regret and
  experimental design.
\newblock In {\em Proc. International Conference on Machine Learning (ICML)},
  2010.

\bibitem{ZiyuWang2013}
Ziyu Wang, Masrour Zoghi, Frank Hutter, David Matheson, and Nando de~Freitas.
\newblock Bayesian optimization in high dimensions via random embeddings.
\newblock In {\em International Joint Conferences on Artificial Intelligence
  (IJCAI) - Distinguished Paper Award - Extended version:
  http://arxiv.org/abs/1301.1942}, 2013.

\end{thebibliography}
}

\end{document}